%% file: main.tex
\documentclass[preprint,12pt,authoryear,nopreprintline]{elsarticle}
\usepackage[T1]{fontenc}
\usepackage[utf8]{inputenc}
\usepackage{lmodern}
\usepackage{microtype}
\usepackage{amsmath}
\usepackage{booktabs}
\usepackage{tabularx}
\usepackage{array}
\usepackage{graphicx}
\usepackage{xcolor}
\usepackage{xurl}
\usepackage{hyperref}
\hypersetup{colorlinks=true,linkcolor=black,urlcolor=blue,citecolor=black}

\newcommand{\spp}{\emph{spp.}}
\definecolor{sdgray}{gray}{0.55}
\newcommand{\sd}[1]{\textcolor{sdgray}{\mbox{\footnotesize\ensuremath{\pm\,#1}}}}
\newcommand{\bestmean}[1]{\textbf{#1}}
\newcommand{\secondmean}[1]{\underline{#1}}
\newcommand{\tablefont}{\fontsize{10}{12}\selectfont}
\newcolumntype{Y}{>{\raggedright\arraybackslash}X}
\journal{Computers and Electronics in Agriculture}
\begin{document}
\begin{frontmatter}
\title{WoodBR-Cross: A public wood-image dataset and individual-tree-disjoint benchmark for species identification}

\author[1]{Daniel Casanova\corref{cor1}}
\ead{danielcasanova@alunos.utfpr.edu.br}
\author[2]{Alexandre Bahia Gontijo}
\author[3]{Arnaldo Candido Junior}

\author[4]{Silvana Nisgoski}
\author[1]{Pedro Luiz de Paula Filho}
\cortext[cor1]{Corresponding author}

\address[1]{Department of Computer Science, Federal University of Technology--Paran\'a (UTFPR), Medianeira, Paran\'a, Brazil}
\address[2]{Forest Products Laboratory (Laborat\'orio de Produtos Florestais), Brazilian Forest Service (Servi\c{c}o Florestal Brasileiro), Bras\'ilia, Brazil}
\address[3]{S\~ao Paulo State University (UNESP), S\~ao Jos\'e do Rio Preto, S\~ao Paulo, Brazil}
\address[4]{Department of Forest Engineering and Technology, Federal University of Paran\'a (UFPR), Paran\'a, Brazil}

\begin{abstract}
Wood identification supports timber inspection and species conservation. However, benchmarks can overestimate generalization when photographs of the same tree occur in training and test sets. We present WoodBR-Cross: 4,875 transverse-surface microscope images, 38 operational taxonomic classes and 303 catalogue-linked trees. A species subset contains 3,406 images, 20 classes and 214 trees. Task A evaluates all classes using duplicate-aware image partitions. B1 partitions images of the species subset, allowing different photographs of one tree across partitions. B2 keeps each tree's photographs together, testing recognition of unseen trees on the same subset. We benchmark three backbones and baselines across three folds and five training seeds. Validation-selected ConvNeXt-Tiny achieves B2 tree-level macro-F1 of $0.8621\pm0.0317$; frozen DINOv2 features yield $0.9000\pm0.0722$. Values are means $\pm$ SD across three folds; fine-tuned scores are seed-averaged within folds. This resource supports reproducible, group-aware evaluation.
\end{abstract}
\end{frontmatter}

\section{Introduction}
\label{sec:introduction}
Wood identification supports timber traceability, quality control and the verification of traded forest products. Once trees have been processed into logs or lumber, many external botanical characters are lost, and identification relies on wood anatomy. Transverse surfaces reveal arrangements of vessels, parenchyma and rays that can help distinguish taxa. Interpreting these structures requires expertise, motivating computer-vision tools that can assist wood inspection \citep{hwang2021,ravindran2020}.

Brazilian research has established important image resources for this task. The microscopic Forest Species Database (FSD) contains 2,240 images from 112 species, spanning 85 genera and 30 families \citep{martins2013}. The public macroscopic FSD-M release contains 2,942 images from 41 species of Brazilian flora \citep{paulafilho2014,fsdm2023}. A separate collection from the Federal University of Paran\'a contains 1,901 transverse-surface images from 46 species and accompanies a texture-based recognition study \citep{souza2020,souzadata2020}. Work using material from the Brazilian Forest Service also examined transverse, radial and tangential images across 281 species \citep{degeus2020}. These resources establish that public Brazilian wood-image datasets predate WoodBR-Cross.

Recognition methods have progressed from handcrafted texture descriptors to learned representations. Early convolutional networks were evaluated on both microscopic and macroscopic Brazilian datasets \citep{hafemann2014}; subsequent studies investigated pretrained networks across anatomical sections \citep{degeus2020}. More recently, \citet{zielinski2025} evaluated deep feature transfer and fusion on a multi-view Congolese wood dataset, while \citet{spancerski2026} compared YOLO architectures using heterogeneous Brazilian macroscopic image sources. These developments make public datasets valuable for comparing both fine-tuned networks and reusable visual features. However, differences in label sets, acquisition conditions and test protocols prevent direct rankings based on published accuracy alone.

The evaluation unit is especially important when several images come from one tree. Different photographs may share tree-specific anatomy or acquisition conditions even when their file contents are distinct. An image-level split can therefore test recognition of a tree already represented in training. Cross-validation for structured observations should preserve the relevant grouping when the intended target is a new group \citep{roberts2017}. In wood identification, \citet{ma2024} showed that image allocation and aggregation of patch predictions affect reported accuracy.

Specimen-aware wood data also exist internationally. IC4SD-Wood-Eucalyptus provides 2,910 images from 86 physical specimens, covering seven \emph{Eucalyptus} species and one outgroup, with specimen identifiers, split manifests and leakage audits \citep{nguyentrong2026}. Conversely, \citet{spancerski2026} report that their Brazilian source datasets lack identifiers linking photographs to physical wood blocks. This identifies a practical need for catalogue-linked image resources and explicit group-based benchmarks, without implying that specimen-aware datasets are unprecedented.

WoodBR-Cross addresses this need with images linked to individual trees through the LPF/SFB collection catalogue. We compare image-based evaluation with individual-tree-disjoint evaluation on a common species-level subset, while separately retaining genus-level operational classes for a broader task. The contribution lies in the documented data and evaluation design rather than a new network architecture. Specifically, we:
\begin{itemize}
  \item release 4,875 images across 38 operational classes, linked to 303 distinct trees by catalogue witness codes, with taxonomic, curation and checksum metadata;
  \item provide duplicate-aware image partitions and B2 partitions that keep every tree wholly within training, validation or test, with saved assignments for reproducible evaluation;
  \item benchmark three pretrained backbones across folds and seeds, alongside handcrafted and frozen-feature baselines and visual-cue controls; and
  \item distinguish image-level errors from tree-level decisions to clarify what each reported result measures.
\end{itemize}

The remainder of this article is organized as follows. Section~\ref{sec:methods} describes the dataset, acquisition documentation and evaluation protocols. Section~\ref{sec:results} presents the benchmark results. Section~\ref{sec:discussion} discusses their implications and limitations, and Section~\ref{sec:conclusions} concludes the article.

\section{Materials and methods}
\label{sec:methods}
\subsection{Dataset provenance, curation and taxonomic labels}
WoodBR-Cross contains microscope photographs of transverse wood surfaces from material linked to the Xiloteca Dr. Harry van der Slooten (FPBw), maintained by the Forest Products Laboratory (LPF), Brazilian Forest Service (SFB), in Bras\'ilia, Brazil \citep{lpf2026}. We visually screened the photographs and excluded images in which wood anatomy could not be distinguished because of severe blur, bark obscuring the surface, or inadequate sanding. All 4,875 retained images were judged to have good visual quality; no numerical blur threshold was used. File-integrity and checksum checks were performed on the retained images. The authors confirmed permission to redistribute the images and associated metadata.

The dataset contains 38 operational classes and 303 distinct trees identified by catalogue witness codes. Twenty classes have species-level metadata (3,406 images, 214 trees); the other 18 are genus-level \spp{} classes (1,469 images). The Xiloteca catalogue assigns a numeric catalogue witness code to each individual tree. We use this code as the tree identifier in B2; the tree identifiers attached to the released images and their catalogue links are consistent across all 4,875 records. All 303 codes map to one catalogue record and barcode in the official export updated on 2 July 2026 \citep{lpf2026}. Nine catalogue codes are associated with two derived grouping keys. To avoid treating one tree as two groups, B2 uses the original catalogue code rather than these derived keys. Consequently, images of one tree cannot cross partitions. The corresponding records can be publicly consulted\footnote{\url{https://lpf.jbrj.gov.br/v2/consulta.php}}.

Each image is associated with metadata organized into 48 fields describing file locations, taxonomic labels, specimen links, curation decisions, visual quality, integrity and SHA-256 checksums. Catalogue family, genus and species fields agree with the released metadata for all images. Of the 303 linked tree records, 281 provide geographic descriptions, 97 name a taxonomic determiner and 276 contain a complete determination date. Other values are absent from the export. The correlated-collection field provides no herbarium-voucher links for this subset. Determination dates, record-access dates and filename timestamps are not interpreted as collection or image-acquisition dates.

Prediction targets follow \texttt{species\_class}; a catalogue species name may coexist with a genus-level operational \spp{} label. Family and subfamily labels follow the supplied metadata and folder hierarchy. We retain these labels without independent botanical reidentification or harmonization against an accepted-name reference. Task A spans five recorded families and three subfamilies, with subfamily values missing for eight classes. It includes one \emph{Acacia} \spp{} class with 193 images from 18 trees. B1/B2 contain 14 genera and three subfamilies, all within Fabaceae. Class support ranges from 21 to 346 images and 2 to 25 trees in A, and from 53 to 346 images and 6 to 25 trees in B1/B2. Median image counts are 102 and 145, respectively. Full counts are supplied in T16 and \texttt{taxonomy.csv}.

The available documentation does not specify collection dates or sampling procedures. It also lacks a detailed surface-preparation protocol, microscope and camera specifications, magnification or physical scale, lighting conditions, imaging location and dates, and the initial and excluded image counts. Together, these gaps limit replication of image acquisition and visual screening.

\begin{figure}[htbp]
\centering
\includegraphics[width=\linewidth]{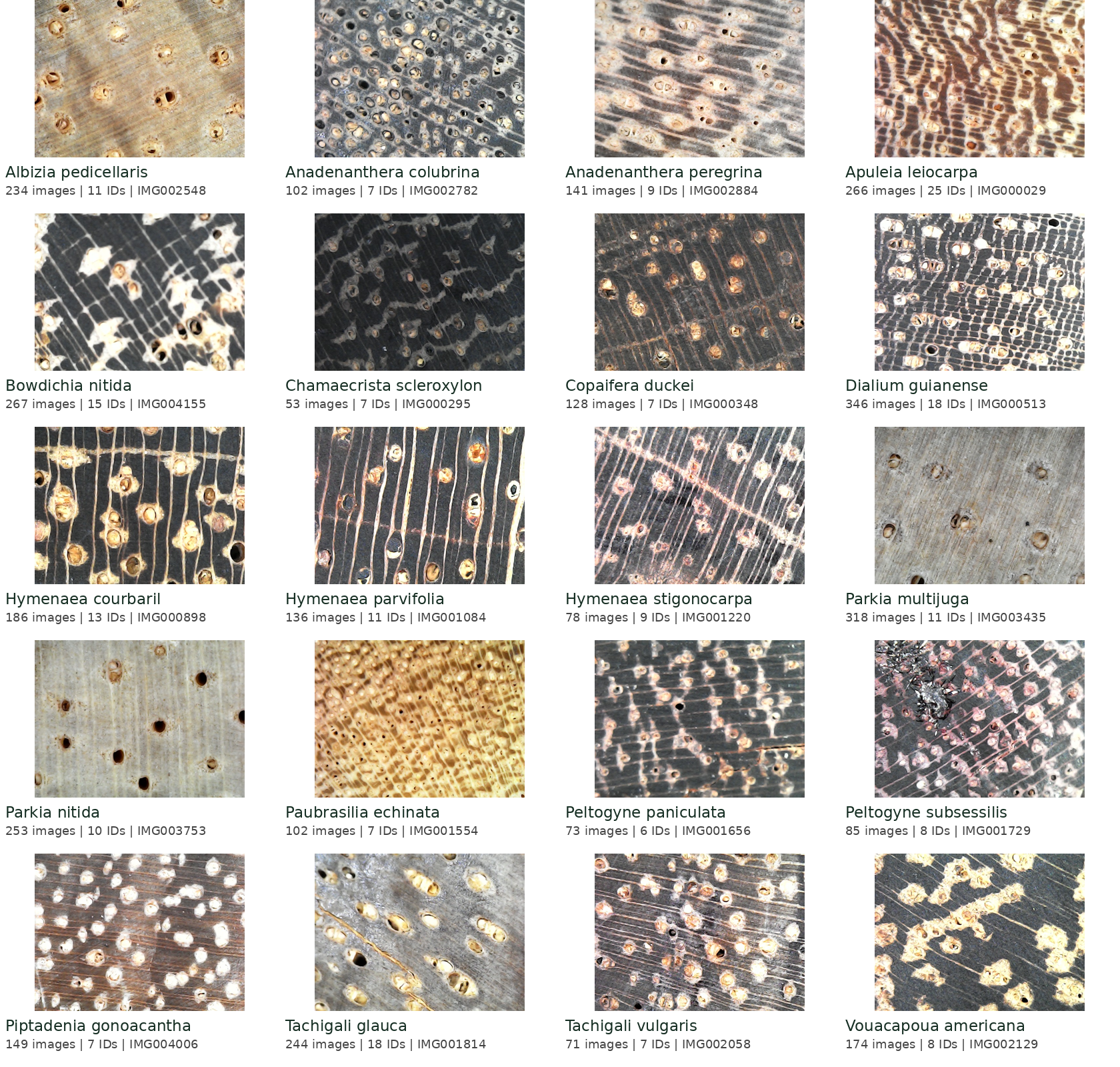}
\caption{Microscope photographs of transverse wood surfaces, with one example per species-level class. Magnification and physical scale are not documented. Labels report image support, distinct-tree support and the example image identifier. Examples were selected by recorded quality and then the smallest image identifier. Files contain no EXIF/GPS information and were not retouched. No physical scale is inferred.}
\label{fig:datasetexamples}
\end{figure}

Nine duplicate-connected components contain two image records each, for 18 records in total. Components were formed from repeated normalized source paths or identical SHA-256 digests. We retained duplicate records but kept all members of each duplicate group within the same partition. One missing source path was resolved to an existing file only after its SHA-256 digest matched. No perceptual search for visually similar files with different paths and hashes was documented. Split manifests verify duplicate-group integrity in all tasks and complete individual-tree separation in B2. Figure~\ref{fig:datasetexamples} illustrates the species-level subset.

\begin{table}[htbp]
\centering\tablefont
\renewcommand{\arraystretch}{1.18}
\setlength{\tabcolsep}{4pt}
\caption{Dataset subsets and evaluation units. Each tree is identified by one LPF catalogue witness code. B1 and B2 use the same 20 species-level classes and the same images; they differ in whether photographs of one tree may cross partitions.}
\label{tab:tasks}
\begin{tabularx}{\linewidth}{@{}lrrrYY@{}}
\toprule
Task & Classes & Images & Trees & Partition unit & Macro-F1 unit \\
\midrule
A & 38 & 4,875 & 303 & Image, with duplicate groups intact & Image \\
B1 & 20 & 3,406 & 214 & Image, with duplicate groups intact & Image \\
B2 & 20 & 3,406 & 214 & Tree ID & Tree \\
\bottomrule
\end{tabularx}
\end{table}

\subsection{Evaluation tasks and data partitions}
We evaluate three tasks that distinguish taxonomic coverage from the unit used to partition observations (Table~\ref{tab:tasks}). Task A includes all 38 operational classes and partitions images while keeping duplicate groups intact. B1 restricts A to the 20 species-level classes and retains each image's original partition assignment. It therefore evaluates image classification without requiring tree separation: different, nonduplicate photographs of one tree may occur in training, validation and test.

B2 uses exactly the same 3,406 images and 20 classes as B1, but partitions the 214 trees instead of their photographs. All images of a tree remain together, so its test images come from trees absent from both training and validation within that fold. For example, B1 permits two different photographs of one tree to be assigned to training and test, whereas B2 requires both photographs to remain in the same partition. B1 and B2 thus address different evaluation scenarios using a common species subset. Their primary metrics also differ: B1 scores individual images, whereas B2 aggregates image probabilities into one decision per tree. We additionally report B2 image-level scores to compare the tasks using the same measurement unit.

All tasks use three outer folds with partition seed 42. For A/B1, the image-level procedure uses \texttt{StratifiedGroupKFold} with duplicate components as groups. B1 preserves the assignments inherited from A rather than generating an independent split. For B2, \texttt{StratifiedKFold} operates on one row per distinct tree ID (the catalogue witness code), and each assignment is expanded to all images from that tree. Both splitters use three folds and shuffling. Internal validation selects the closest feasible 20\% per class while keeping duplicate groups intact in A/B1 and tree IDs intact in B2. Within each B2 fold, tree IDs are mutually exclusive across training, validation and test; no images from one tree occur in more than one partition. This prevents individual-tree leakage: test-tree images do not enter training or model selection. Each tree appears in one outer-test fold per training seed; its role rotates across the three cross-validation folds. Table~\ref{tab:splitcounts} gives partition totals; supplementary manifests provide class-by-fold counts and tree assignments.

\begin{table}[htbp]
\centering\tablefont
\renewcommand{\arraystretch}{1.15}
\setlength{\tabcolsep}{8pt}
\caption{Partition sizes. B2 counts distinct trees identified by catalogue witness code; tree IDs are disjoint across training, validation and test within each fold. Training and validation are combined for final fitting after model selection.}
\label{tab:splitcounts}
\textbf{(a) Images}\par\smallskip
\begin{tabular}{@{}lrrrr@{}}
\toprule
Task & Fold & Training & Validation & Test \\
\midrule
A & 1 & 2,600 & 649 & 1,626 \\
A & 2 & 2,600 & 649 & 1,626 \\
A & 3 & 2,601 & 651 & 1,623 \\
\addlinespace
B1 & 1 & 1,817 & 454 & 1,135 \\
B1 & 2 & 1,817 & 454 & 1,135 \\
B1 & 3 & 1,816 & 454 & 1,136 \\
\addlinespace
B2 & 1 & 1,766 & 442 & 1,198 \\
B2 & 2 & 1,910 & 468 & 1,028 \\
B2 & 3 & 1,717 & 509 & 1,180 \\
\bottomrule
\end{tabular}
\par\medskip
\textbf{(b) Distinct trees in B2}\par\smallskip
\begin{tabular}{@{}rrrr@{}}
\toprule
Fold & Training & Validation & Test \\
\midrule
1 & 115 & 27 & 72 \\
2 & 117 & 26 & 71 \\
3 & 115 & 28 & 71 \\
\bottomrule
\end{tabular}
\end{table}

\subsection{Backbones, model selection and training}
The backbones are ResNet-50 \citep{he2016}, ConvNeXt-Tiny \citep{liu2022} and ViT-B/16 \citep{dosovitskiy2021}. Their \texttt{timm} checkpoints are \path{resnet50.a1_in1k}, \path{convnext_tiny.in12k_ft_in1k} and \path{vit_base_patch16_224.augreg2_in21k_ft_in1k}, respectively. Pretraining sources differ, so comparisons concern these pretrained configurations rather than architecture alone.

For each task, fold and backbone, a six-point grid crosses learning rates $\{10^{-5},3\times10^{-5},10^{-4}\}$ with weight decay $\{0.01,0.05\}$, giving 162 selection fits. Image-level macro-F1 selects configurations in A/B1; tree-level macro-F1 does so in B2. A/B1 ties follow grid order and then earlier epoch. B2 ties use image-level macro-F1, validation loss and then earlier epoch. The average selected validation score across folds determines the task-level backbone: ViT-B/16 for A/B1 and ConvNeXt-Tiny for B2. Test scores do not enter this choice.

Each backbone is then fitted on training plus validation data for its selected epoch count, without test-based early stopping. Five training seeds (42, 123, 2024, 7 and 21) give 135 final fits. Fold-specific configurations are archived in T13; task-level selections are recorded in \texttt{verified\_validation\_selection.csv}.

Inputs are $224\times224$ pixels. Training uses random resized cropping (scale 0.5--1.0; aspect ratio 0.9--1.1), horizontal flipping ($p=0.5$), and brightness/contrast jitter of 0.1, without saturation or hue jitter. Evaluation resizes the shorter side to 256 pixels and center-crops to $224\times224$ using bicubic interpolation. ResNet-50 and ConvNeXt-Tiny use channel means $(0.485,0.456,0.406)$ and standard deviations $(0.229,0.224,0.225)$; ViT-B/16 uses 0.5 for both.

Training uses AdamW, batch size 32 and CUDA mixed precision with fp16 autocast and GradScaler. Image-level cross-entropy uses inverse image-frequency class weights normalized to a mean of one. No tree-level weighting is added, so trees contributing more images have greater aggregate influence during B2 training. Selection permits up to 30 epochs with patience five. The learning rate warms from 10\% of the target during the first epoch and then follows cosine decay. Seeds are set, but cuDNN benchmarking and TF32 are enabled; numerical reproduction bit for bit is therefore not claimed.

Recorded durations total 18,763.5~s (5.21~h) across 373 timed evaluations. Baselines and two controls have no recorded durations, so this total is a partial sum of run timings rather than total campaign wall time or verified GPU-hours.

\subsection{Baselines and visual-cue controls}
Baselines reuse the saved outer and internal partitions. Candidates are selected on internal validation and refitted on training plus validation before external testing. We evaluate a majority-class predictor, handcrafted color/texture descriptors with an RBF support vector classifier (SVC) or random forest, and frozen DINOv2 features \citep{oquab2023} with logistic regression or $k$-nearest neighbors ($k$NN).

Handcrafted features combine HSV color histograms, local binary patterns (LBP) and gray-level co-occurrence matrices (GLCM), capturing complementary color and texture information. Table~\ref{tab:baselineconfig} summarizes the descriptor settings and classifier candidates.

\begin{table}[htbp]
\centering\tablefont
\renewcommand{\arraystretch}{1.12}
\caption{Handcrafted descriptor settings and classifier selection grid. Candidate classifier settings are selected using internal validation.}
\label{tab:baselineconfig}
\begin{tabularx}{\linewidth}{@{}lY@{}}
\toprule
Parameter & Value \\
\midrule
HSV histogram & 16 bins per channel; each channel normalized separately \\
Uniform LBP & $P=16$ neighbors; radius $R=2$ \\
LBP/GLCM input & Grayscale, resized to $128\times128$ pixels \\
GLCM quantization & 16 gray levels \\
GLCM offsets & Distances 1, 2 and 4; four angles \\
GLCM statistics & Contrast, dissimilarity, homogeneity, angular second moment, energy and correlation \\
RBF SVC & $C\in\{0.1,1,10\}$ \\
Random forest size & 400 decision trees \\
Random forest feature sampling & Square root of feature count or 0.5 of features \\
Random forest minimum leaf size & 1 or 3 samples \\
\bottomrule
\end{tabularx}
\end{table}

Frozen features come from \path{vit_small_patch14_dinov2.lvd142m} and are $L_2$-normalized. Logistic regression tests $C=0.01,0.1,1,10$; $k$NN tests $k=1,3,5,9$ with uniform or distance weighting. SVC and logistic-regression inputs are standardized. The SVC uses its built-in probability estimator; no separate post-hoc calibration is applied.

Two controls probe available visual cues. The border-only control retains the outer 15\% strip on each side and replaces the center with gray value 127 after transformation, during training, validation and testing. The randomized-label control permutes labels globally among duplicate components in A/B1 and among tree IDs in B2, rather than within class strata. Both use the task-level validation-selected backbone and the corresponding test protocol. These controls do not isolate anatomical structures from background or acquisition cues.

\subsection{Metrics and statistical summaries}
Macro-F1 is the primary metric because classes have unequal support. A/B1 evaluate each image. B2 averages all image probabilities from each tree before assigning one class, giving one equally weighted decision per distinct tree ID. Accuracy, balanced accuracy, top-3 accuracy, expected calibration error (ECE; 15 bins) and multiclass Brier score are secondary image-level metrics, including in B2.

For the fine-tuned backbones, five seed scores are first averaged within each fold. We report the mean and sample standard deviation (SD) of the three resulting fold means. The 15 fold--seed runs share partitions and are not independent external replications. Baseline/control tables summarize the three available fold scores. These SDs describe fold variability; they are not confidence intervals. We report no bootstrap intervals or paired inferential comparisons.

Confusion matrices use image-level out-of-fold (OOF) predictions in every task. Each seed's matrix is normalized by true class, and the five matrices are averaged. Consequently, the B2 matrix describes images from held-out trees, whereas primary B2 macro-F1 describes tree-level decisions aggregated over all images from each tree.

\subsection{Representation analysis and learning curves}
To represent a tree using several photographs, we first $L_2$-normalize each image embedding, average these unit vectors within the tree, and normalize the resulting mean again. Normalizing before averaging gives each photograph equal weight in the mean, rather than allowing embeddings with larger norms to contribute more strongly. The second normalization expresses the tree representation as a unit vector. Although $L_2$ normalization does not change the cosine similarity between two individual vectors, normalizing each image before aggregation can change the direction of the tree mean. This step therefore specifies how multiple images contribute to a tree representation; we do not claim a performance gain from normalization without an ablation. Pairs of distinct B2 test trees within a fold are assigned their closest recorded taxonomic relation: same family/different subfamily, same subfamily/different genus, same genus/different class, or same class. The four groups contain 3,635, 3,301, 234 and 356 distinct pairs, respectively; these pairs are repeated across seeds. Weighted Spearman trends are assessed with 1,000 complete-profile permutations. Because B2 contains only Fabaceae, it cannot assess between-family similarity.

Supplementary t-SNE projections summarize the 214 OOF trees. Trained and frozen representations are projected separately after PCA, using PCA initialization, automatic learning rate, perplexity 30 and seed 42. Grad-CAM \citep{selvaraju2017} illustrates high-confidence correct and incorrect examples for the selected models in fold 1/seed 42. Historical checkpoints were unavailable, so weights were reconstructed: prediction labels agreed, and maximum probability discrepancies were 0 in A/B1 and 0.00012 in B2. The additional B2 error examples presented below use fold 1/seed 123 and a separate reconstruction checked against archived predicted labels. These analyses describe representations and image regions; they do not establish phylogeny or causal anatomical reliance.

The B2 learning curve holds the outer-test trees fixed and samples $k=1,2,3,4$ training trees per class, retaining all their images. Five subsets per $k$ in each of three folds give 60 runs. Within-fold validation selects ResNet-50, ViT-B/16 and ConvNeXt-Tiny in folds 1, 2 and 3, respectively. Configurations and sampled codes are archived. In the learning-curve plot, horizontal offsets separate fold-specific point groups and individual runs for readability; all points around an integer $k$ use exactly $k$ training trees per class. These offsets do not encode training-image counts or an architectural ranking. Each $k$ is summarized by the mean and descriptive SD of 15 fold-by-subset runs. Increasing $k$ increases both the number of trees and the number of images, so their effects cannot be separated by this design.

\section{Results}
\label{sec:results}
\subsection{Image-based and individual-tree-disjoint classification}
\begin{figure}[!htbp]
\centering
\includegraphics[width=\linewidth]{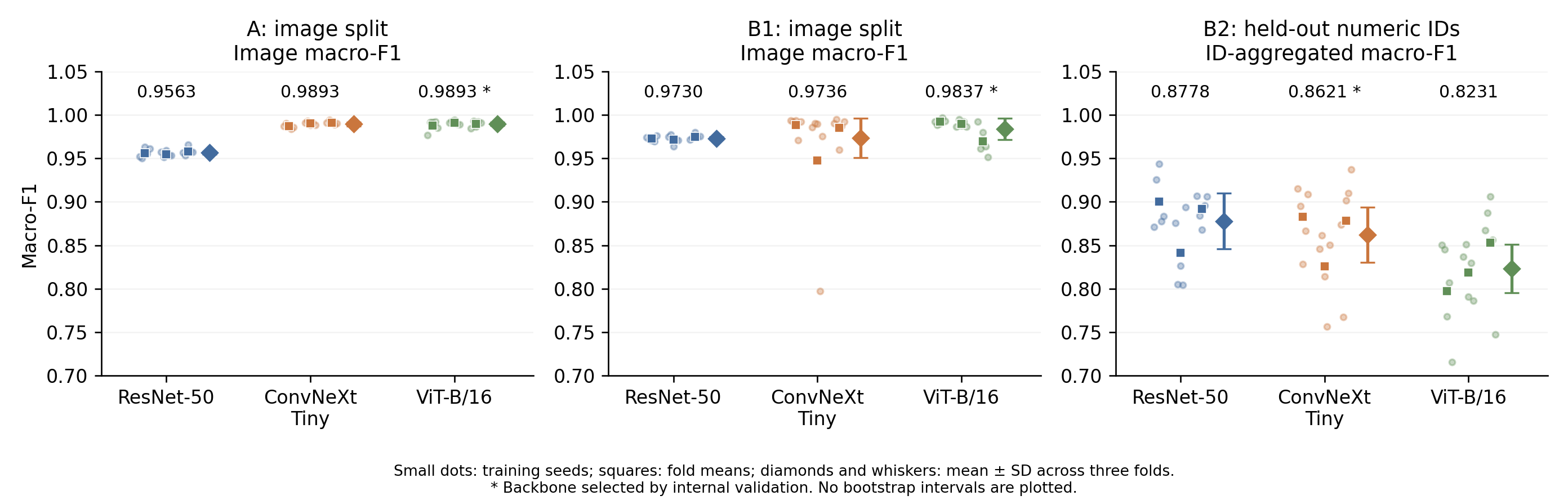}
\caption{Primary macro-F1 across folds and training seeds. Dots are seed scores, squares are fold means, and diamonds with whiskers are the mean and SD across three folds. A/B1 evaluate images; B2 evaluates distinct trees (numeric catalogue witness codes in the figure). Asterisks mark validation-selected backbones.}
\label{fig:primary}
\end{figure}

Validation-selected ViT-B/16 achieved image-level macro-F1 of $0.9893\pm0.0019$ in A and $0.9837\pm0.0123$ in B1. In the individual-tree-disjoint task B2, selected ConvNeXt-Tiny achieved tree-level macro-F1 of $0.8621\pm0.0317$ (Table~\ref{tab:performance}; Figure~\ref{fig:primary}). ResNet-50 had a higher B2 test mean of $0.8778\pm0.0321$, but ConvNeXt-Tiny remained the validation-selected model. This distinction preserves separation between model selection and external assessment.

\input{tables/performance_readable.tex}

Comparing primary B1 and B2 scores changes both the partition unit and the measurement unit. Table~\ref{tab:sameunit} therefore also compares the two tasks at image level. For ConvNeXt-Tiny, image-level macro-F1 decreased from $0.9736\pm0.0226$ in B1 to $0.7979\pm0.0260$ in B2, a difference of approximately 0.176. This comparison keeps the class set and metric unit fixed, but task-specific configurations and test composition still differ. It is a descriptive comparison of evaluation scenarios rather than a causal estimate of the effect of splitting.

\begin{table}[!htbp]
\centering\tablefont
\renewcommand{\arraystretch}{1.15}
\setlength{\tabcolsep}{4pt}
\caption{Macro-F1 for B1 and B2 by evaluation unit. Values are means $\pm$ SD across three seed-averaged fold means. Bold marks the best displayed mean; underlining marks the next-highest distinct mean in each column.}
\label{tab:sameunit}
\begin{tabular}{@{}lccc@{}}
\toprule
Backbone & \shortstack{B1\\image} & \shortstack{B2\\image} & \shortstack{B2\\tree-level} \\
\midrule
ResNet-50 & 0.9730\sd{0.0017} & \secondmean{0.7749}\sd{0.0185} & \bestmean{0.8778}\sd{0.0321} \\
ConvNeXt-Tiny & \secondmean{0.9736}\sd{0.0226} & \bestmean{0.7979}\sd{0.0260} & \secondmean{0.8621}\sd{0.0317} \\
ViT-B/16 & \bestmean{0.9837}\sd{0.0123} & 0.7583\sd{0.0293} & 0.8231\sd{0.0279} \\
\bottomrule
\end{tabular}
\end{table}

\subsection{Baseline and visual-cue control performance}
In B2, frozen DINOv2 features yielded tree-level macro-F1 of $0.9000\pm0.0722$, compared with $0.7024\pm0.0952$ for handcrafted descriptors and $0.0069\pm0.0014$ for the majority predictor (Table~\ref{tab:baselines}). The frozen-feature mean exceeds that of the validation-selected fine-tuned backbone, but no paired inferential comparison establishes superiority. The results do not demonstrate that end-to-end fine-tuning is necessary for this collection.

\begin{table}[htbp]
\centering\tablefont
\renewcommand{\arraystretch}{1.2}
\setlength{\tabcolsep}{4pt}
\caption{Baseline and control macro-F1. Values are means $\pm$ SD of three fold scores. Among baselines, bold and underlining mark the best and second-best mean in each task; controls are unranked.}
\label{tab:baselines}
\begin{tabularx}{\linewidth}{@{}Xccc@{}}
\toprule
Method & \shortstack{A\\images} & \shortstack{B1\\images} & \shortstack{B2\\trees} \\
\midrule
\multicolumn{4}{@{}l}{\emph{Baselines}} \\
HSV + LBP + GLCM & \secondmean{0.9221}\sd{0.0126} & \secondmean{0.9527}\sd{0.0046} & \secondmean{0.7024}\sd{0.0952} \\
Frozen DINOv2 & \bestmean{0.9475}\sd{0.0154} & \bestmean{0.9719}\sd{0.0070} & \bestmean{0.9000}\sd{0.0722} \\
Majority class & $0.0035\sd{0.0000}$ & $0.0092\sd{0.0000}$ & $0.0069\sd{0.0014}$ \\
\addlinespace
\multicolumn{4}{@{}l}{\emph{Controls}} \\
Border-only inputs & $0.9804\sd{0.0093}$ & $0.9726\sd{0.0181}$ & $0.8428\sd{0.0798}$ \\
Randomized labels & $0.0181\sd{0.0026}$ & $0.0277\sd{0.0150}$ & $0.0344\sd{0.0324}$ \\
\bottomrule
\end{tabularx}
\end{table}

Border-only inputs retained B2 tree-level macro-F1 of $0.8428\pm0.0798$, whereas randomized labels yielded $0.0344\pm0.0324$. Information in the image periphery thus remained discriminative. The border mask retains both peripheral wood anatomy and possible contextual information, so its performance does not identify which source explains the predictions. Center-only and background-segmented evaluations were not performed in the reported comparison.

\subsection{Image-level errors on held-out trees}
\begin{figure}[!htbp]
\centering
\includegraphics[width=0.78\linewidth]{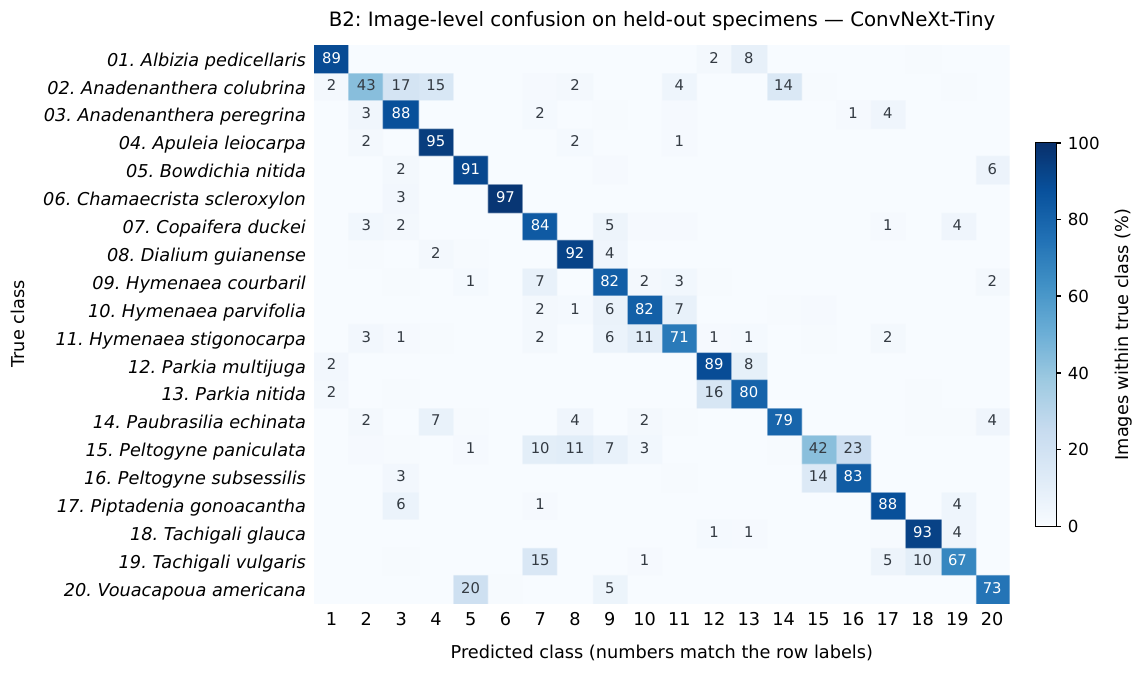}
\caption{Image-level OOF confusion for ConvNeXt-Tiny in B2: mean of five seed matrices normalized by true class. Column numbers match the numbered row labels. Annotations are rounded percentages; cells below 1\% are unannotated. Images come from held-out trees; predictions are not aggregated by tree in this matrix. Class support is reported in T16.}
\label{fig:confusionB2}
\end{figure}

Figure~\ref{fig:confusionB2} reports the image-level OOF confusion matrix for ConvNeXt-Tiny, selected by validation in B2. Errors were concentrated in a subset of species. The five classes with the highest total error rates were \emph{Peltogyne paniculata} (57.5\%), \emph{Anadenanthera colubrina} (57.1\%), \emph{Tachigali vulgaris} (33.2\%), \emph{Hymenaea stigonocarpa} (28.7\%) and \emph{Vouacapoua americana} (26.6\%). Their most frequent incorrect predictions were, respectively, \emph{P.\ subsessilis} (23.0\%), \emph{A.\ peregrina} (17.5\%), \emph{Copaifera duckei} (14.9\%), \emph{H.\ parvifolia} (10.5\%) and \emph{Bowdichia nitida} (20.5\%). Each confusion percentage uses all test images of the true class as its denominator, rather than only its errors. These are image-level rates averaged across five OOF seed matrices, not proportions of misclassified trees. Full class metrics and support are supplied in T08/T16.

\begin{figure}[!htbp]
\centering
\includegraphics[width=\linewidth]{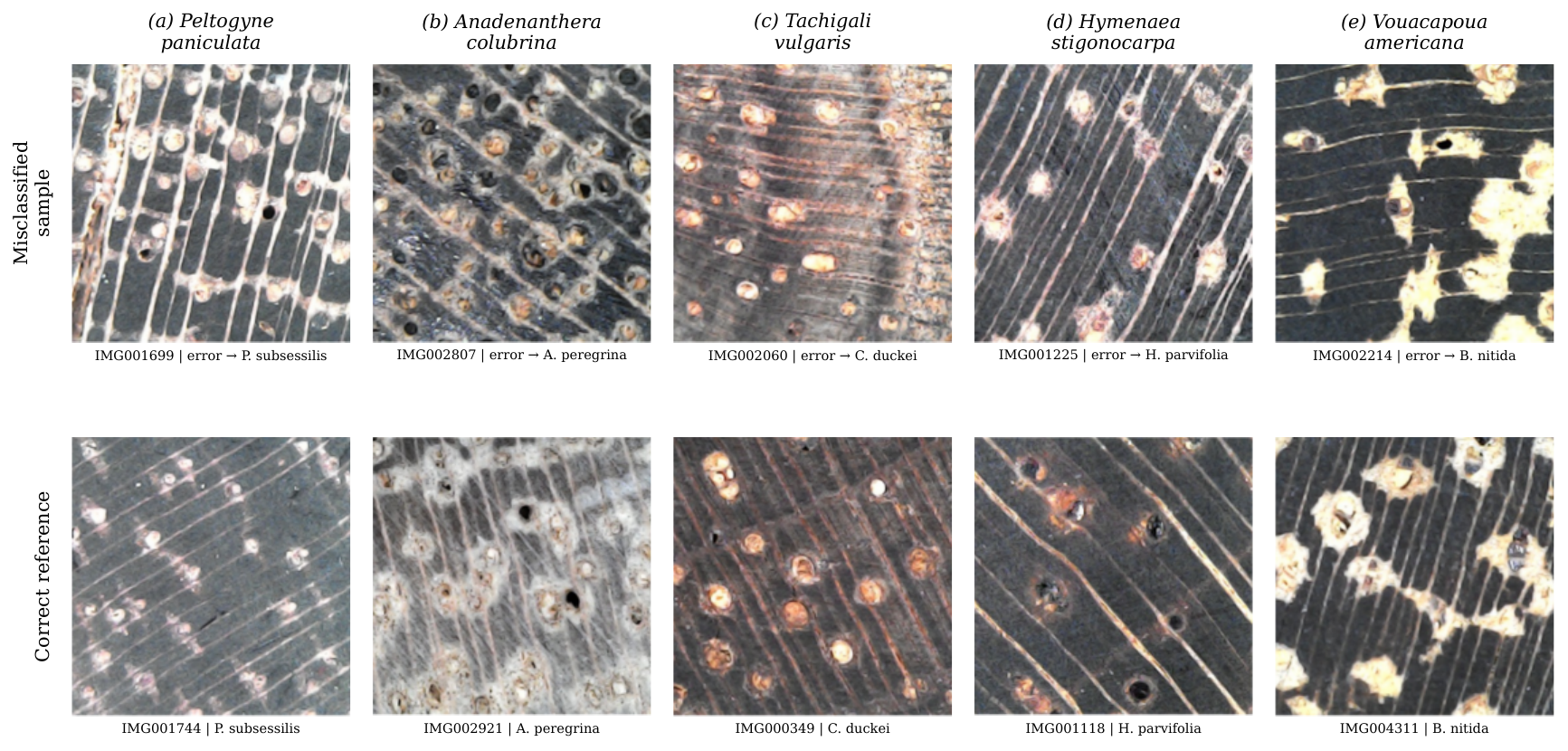}
\caption{Illustrative B2 errors for validation-selected ConvNeXt-Tiny. Columns (a)--(e) follow the five true classes with the highest total OOF error rates. Top: a misclassified test image. Bottom: a correctly classified reference of the species predicted for the image above. Examples come from test fold 1, seed 123. Image identifiers allow the examples to be traced. The references are illustrative examples, not estimates of within-species variation.}
\label{fig:errorPairs}
\end{figure}

Figure~\ref{fig:errorPairs} illustrates these five confusions using one misclassified image and one correctly classified reference from the predicted species. Some examples share visible pore arrangements, linear patterns and contrast, particularly the two \emph{Peltogyne} species and the two \emph{Hymenaea} species. A single example per pair does not establish systematic anatomical similarity.

\begin{figure}[!htbp]
\centering
\includegraphics[width=\linewidth]{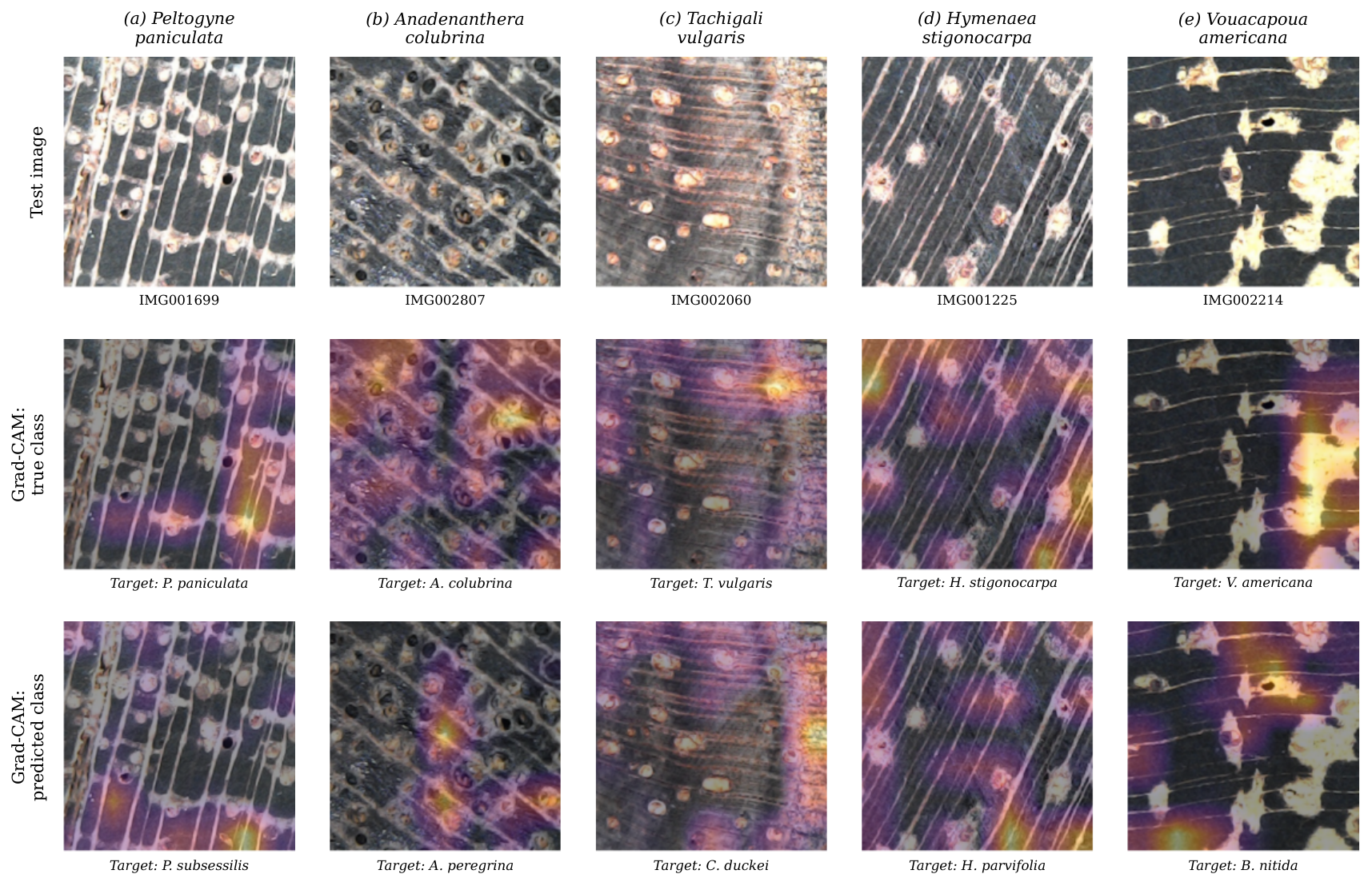}
\caption{Grad-CAM for the misclassified images in Figure~\ref{fig:errorPairs}, using ConvNeXt-Tiny from B2 test fold 1, seed 123. Rows show the original image, attribution for the true class and attribution for the predicted class. Maps are normalized independently, with warmer colors indicating higher activation within each map. Colors therefore do not provide a common quantitative scale across targets or images. These examples do not establish causal reliance on a particular anatomical character.}
\label{fig:errorGradcam}
\end{figure}

The same test images are examined with class-targeted Grad-CAM in Figure~\ref{fig:errorGradcam}. The reconstructed model reproduced 100\% of the archived predicted class labels for this run, as reported in the supplied analysis. This agreement concerns predicted labels and does not establish identical probabilities or attribution maps. The maps show regions contributing to each target-class score and are interpreted qualitatively.

\subsection{Taxonomic associations and training-set size}
Among incorrect B2 tree-level decisions, 41.7--55.7\% remained within genus and 76.4--83.5\% within subfamily across backbones. The corresponding uniform incorrect-class references were 4.8--5.5\% and 50.4--55.2\%; prediction-frequency references were 5.3--6.1\% and 51.6--56.1\%. These counts pool repeated seed decisions rather than additional distinct trees. Family agreement is automatic because all B2 classes belong to Fabaceae.

Taxonomic proximity and embedding similarity were positively associated for trained backbones (weighted $\rho=0.3457$--0.3952; one-sided permutation $p=0.0030$--0.0190). For selected ConvNeXt-Tiny, $\rho=0.3640$ and $p=0.0060$. The frozen reference yielded $\rho=0.2261$ and did not reach the permutation threshold in the corresponding comparisons. These associations describe the recorded label hierarchy and are exploratory; they do not reconstruct phylogenetic relationships.

\begin{figure}[htbp]
\centering
\includegraphics[width=0.94\linewidth]{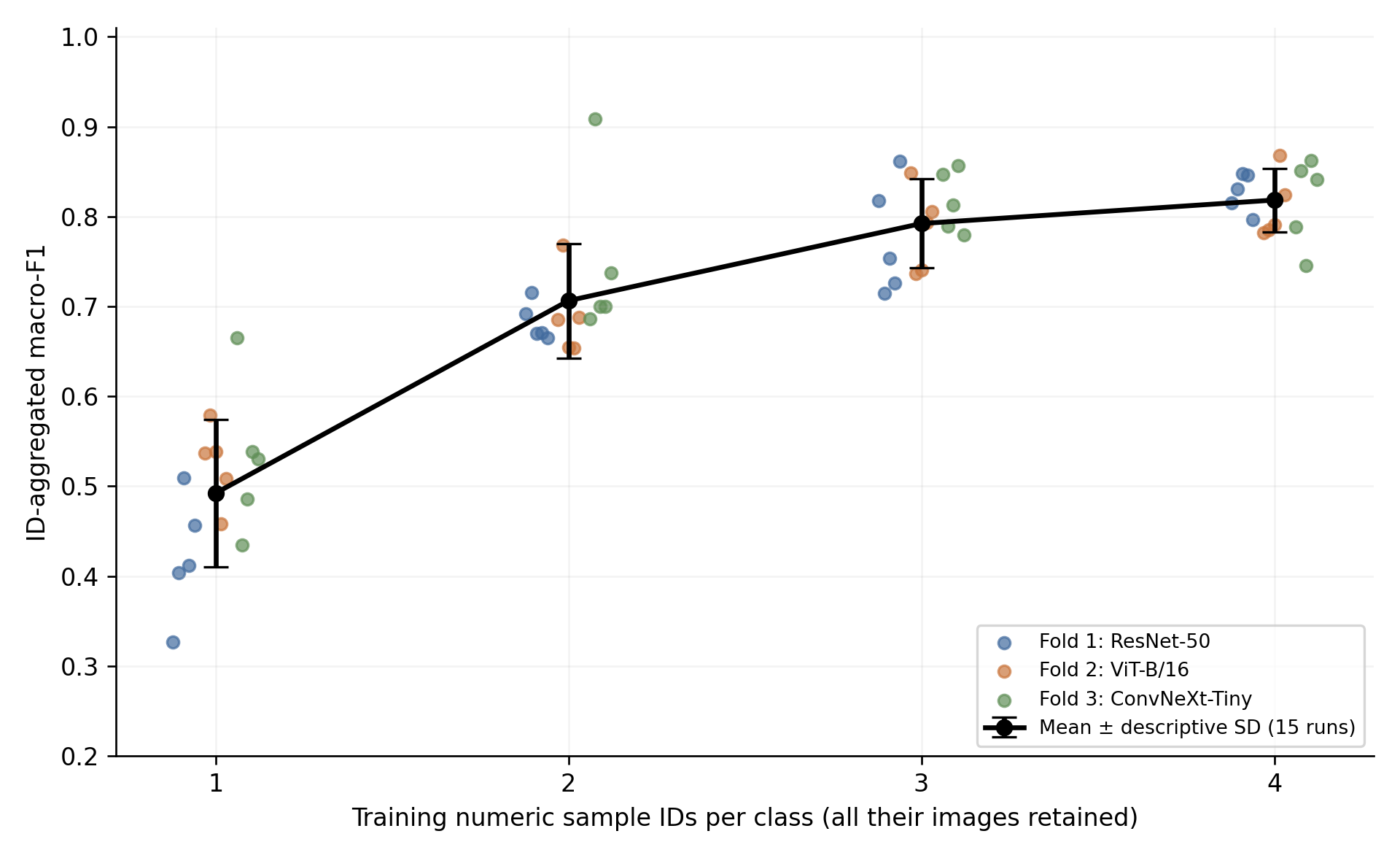}
\caption{The 60 B2 learning-curve runs. The x-axis gives $k$, the number of training trees per class; each $k$ uses five subsets in each of three folds, retaining all images of every sampled tree. Colored points are offset horizontally by fold and within each group solely for readability (jitter); their left-to-right order does not indicate different numbers of training trees or images. ResNet-50, ViT-B/16 and ConvNeXt-Tiny are selected in folds 1, 2 and 3, respectively, so colors also identify folds and do not provide a controlled architecture comparison. Black markers are positioned at the true integer $k$ and show the mean and descriptive SD of 15 runs, not confidence intervals.}
\label{fig:learning}
\end{figure}

Tree-level learning-curve macro-F1 increased from $0.4923\pm0.0816$ with one training tree per class to $0.8183\pm0.0355$ with four (Figure~\ref{fig:learning}). Mean training image counts increased from 304.0 to 1,281.7. This trend combines the number of training trees and images under fold-specific model selection.

\section{Discussion}
\label{sec:discussion}
\subsection{What the benchmark measures}
The evaluation distinguishes two practical questions: recognizing another photograph from a tree already represented in training, and recognizing photographs from a previously unseen tree. The image-level comparison between B1 and B2 shows lower performance when complete trees are held out. In particular, duplicate grouping alone cannot address this difference because photographs of the same tree need not have identical file contents. Tree IDs therefore support a more relevant estimate when future observations are expected to come from new trees, consistent with validation principles for grouped observations \citep{roberts2017}.

Published wood-image accuracies should be interpreted in the context of these units. Brazilian microscopic and macroscopic resources have supported both handcrafted and learned recognition methods \citep{martins2013,paulafilho2014,hafemann2014,souza2020,degeus2020}. Their class coverage and acquisition protocols differ from WoodBR-Cross. Recent multi-source evaluations also identify the lack of physical-specimen identifiers as a constraint \citep{spancerski2026}. We therefore position WoodBR-Cross as a complementary catalogue-linked benchmark rather than claim an overall performance ranking or priority as the first Brazilian dataset. IC4SD-Wood-Eucalyptus likewise demonstrates that specimen identifiers and leakage-audit materials can be released with wood images \citep{nguyentrong2026}.

Beyond the partition design, aggregation answers a different question from image classification. Averaging probabilities gives one B2 decision per tree and equally weights trees at evaluation, whereas the confusion matrix records errors among their images. The difference between these scores is expected and should not be interpreted as contradictory evidence. Related work also shows that aggregation and image allocation can alter wood-recognition accuracy \citep{ma2024}. In our benchmark, image-weighted training still gives trees with more photographs greater influence, even though tree-level evaluation treats them equally.

\subsection{Implications of baselines and controls}
The strong frozen DINOv2 baseline indicates that reusable visual representations are a competitive option for this collection. This agrees with the broader motivation for visual feature transfer \citep{oquab2023} and with wood-identification studies that evaluate transferred features across anatomical sections \citep{zielinski2025}. The comparison here does not establish a universal advantage for frozen features: fold composition, pretraining and downstream selection all contribute to the observed results.

The controls complement these baseline comparisons: high border-only scores require caution when interpreting visual explanations. Peripheral regions still contain wood structure, so a border control cannot separate anatomical signal from acquisition or background cues. Grad-CAM examples and taxonomic associations provide complementary descriptions, but neither demonstrates that the classifier uses the same diagnostic characters as a wood anatomist. Further interventions and expert assessment would be needed to resolve that question.

\subsection{Interpreting recurrent species confusions}
The recurrent errors include both species from the same genus and species from different genera. This pattern is compatible with a fine-grained discrimination task in which taxonomic proximity alone does not account for every confusion. Previous work describes overlapping anatomical patterns and the limits of species-level discrimination from transverse wood images \citep{ravindran2018}. Microscopic classification of Brazilian commercial woods also identifies anatomical similarity and wood variability as practical challenges \citep{moulin2022}. These findings motivate possible explanations for the B2 errors but do not confirm their causes in WoodBR-Cross.

Because B2 separates complete individuals between training and test sets, within-species variation among trees and the limited information in a single transverse photograph may contribute to the observed errors. The examples in Figures~\ref{fig:errorPairs} and~\ref{fig:errorGradcam} cannot distinguish these explanations from acquisition effects or other model cues. Confirmation would require expert assessment of multiple images per species pair and, where available, complementary anatomical views.

\subsection{Scope and limitations}

The reported B2 results concern trees held out by the Xiloteca catalogue witness code. Each code identifies a distinct tree, and every image carrying the same code is assigned to a single partition. Therefore, within each fold, B2 has no individual-tree leakage across training, validation and test; test trees are unseen during model fitting and selection. Additional training seeds characterize optimization variability but do not increase the number of distinct trees in the dataset.

The 20-class subset is restricted to Fabaceae, and operational genus-level labels in A differ in taxonomic resolution from the species-level targets. These aspects should be considered when interpreting and reusing the dataset and its benchmarks. Catalogue-linked metadata and explicit tree-level partitions provide a controlled basis for evaluation within the proposed collection \citep{hwang2021}.

\section{Conclusions}
\label{sec:conclusions}
WoodBR-Cross provides 4,875 wood images, 38 operational classes and catalogue links for 303 distinct trees. Its evaluation protocols separate duplicate-aware image testing from B2 testing on held-out trees. On the 20-class species subset, validation-selected ConvNeXt-Tiny achieved tree-level macro-F1 of $0.8621\pm0.0317$. The lower image-level performance under individual-tree-disjoint partitions, together with strong frozen-feature and border-only results, shows why scores must be interpreted alongside their grouping and measurement units. The central contribution is the public dataset, documented metadata and saved evaluation partitions, including B2 manifests that prevent images from the same tree crossing partitions. These resources support transparent comparisons while external acquisition-domain performance remains to be evaluated.

\section*{Data and code availability}
Images and metadata are publicly available at \url{https://huggingface.co/datasets/Casanovafeat/WoodBR-Cross/tree/main} under Creative Commons Attribution--NonCommercial 4.0 International (CC BY-NC 4.0), \url{https://creativecommons.org/licenses/by-nc/4.0/}. The evaluated release is revision \path{159406cbdbae4b91c30980ccdeb13ade39122f1d}, available at \url{https://huggingface.co/datasets/Casanovafeat/WoodBR-Cross/tree/159406cbdbae4b91c30980ccdeb13ade39122f1d}. The checksums, labels and catalogue codes of its 4,875 images match those of the evaluated collection. The underlying LPF catalogue is separately distributed under the source portal's Creative Commons Attribution terms \citep{lpf2026}.

The repository provides JPEG images and three metadata tables, each containing 4,875 records. The manifest \texttt{metadata.csv} supports the Dataset Viewer. Detailed metadata are provided in \texttt{metadata\_full.csv}. The published quality field uses the Portuguese label \texttt{boas} (good) in \path{quality_status}. Two matching metadata fields store the tree identifier and original catalogue witness code: \texttt{individual\_id} and \path{csv_codtestemunho}. The file \texttt{metadata\_\allowbreak without\_\allowbreak sp.csv} contains an alternate table, not the 20-class experimental subset. The Hugging Face Viewer exposes the complete collection as one \texttt{train} loading split. The optional \texttt{split\_suggested} field in the CSVs does not define this article's cross-validation or prescribe downstream splits. Saved evaluation partitions, configurations and result summaries accompany the article as supplementary files. Training and analysis scripts are not included in the cited dataset-repository revision.

\section*{Funding and acknowledgements}
We thank CNPq for granting the Productivity Fellowship in Technological Development and Innovative Extension (DT-II, Process No.\ 303279/2023-0), which contributed to the completion of this work.
\section*{Declaration of competing interests}
The authors declare that they have no known competing financial interests or personal relationships that could have appeared to influence the work reported in this paper.

\section*{Declaration of generative AI and AI-assisted technologies in the manuscript preparation process}
During the preparation of this manuscript, the authors used ChatGPT (OpenAI) to polish the text and improve its language, readability, and spelling. After using this tool, the authors reviewed and edited the text as needed and take full responsibility for the content of the final manuscript.

\section{Supplementary materials}
Supplementary tables provide dataset counts and taxonomic metadata, split manifests, model-selection and run summaries, class-level error analyses, representation results, and the B2 learning-curve runs. Detailed supporting tables, configurations, and split manifests are provided as supplementary files. In these files, ``individual'' and ``tree'' refer to a distinct tree identified by the Xiloteca catalogue witness code (\emph{c\'odigo testemunho}); B2 keeps all images from each tree in one partition, with no tree-ID overlap among training, validation and test within a fold. Legacy unstratified bootstrap intervals are retained only for traceability and are not reported as valid confidence intervals in this article.

\bibliographystyle{elsarticle-harv}
\begingroup
\tablefont
\bibliography{references}
\endgroup
\end{document}

%% file: tables/performance_readable.tex
\begin{table}[!htbp]
\centering\tablefont
\renewcommand{\arraystretch}{1.22}
\setlength{\tabcolsep}{4pt}
\caption{Outer-test means $\pm$ SD across three seed-averaged fold means. In panel (a), B2 macro-F1 is tree-level, computed after averaging image probabilities within each catalogue witness code; all other metrics in panels (a) and (b) are image-level. Within each task and metric, bold marks all best displayed means (including ties) and underlining marks the next-highest distinct mean. Lower ECE and Brier values are better. Asterisks mark validation-selected backbones.}
\label{tab:performance}
\textbf{(a) Classification}\par\smallskip
\begin{tabular}{@{}llccc@{}}
\toprule
Task & Backbone & Macro-F1 & Accuracy & \shortstack{Balanced\\accuracy} \\
\midrule
A & ResNet-50 & \secondmean{0.9563}\sd{0.0016} & 0.9561\sd{0.0015} & 0.9609\sd{0.0013} \\
A & ConvNeXt-Tiny & \bestmean{0.9893}\sd{0.0023} & \bestmean{0.9914}\sd{0.0027} & \secondmean{0.9877}\sd{0.0023} \\
A & ViT-B/16* & \bestmean{0.9893}\sd{0.0019} & \secondmean{0.9897}\sd{0.0033} & \bestmean{0.9894}\sd{0.0005} \\
\midrule
B1 & ResNet-50 & 0.9730\sd{0.0017} & 0.9756\sd{0.0014} & \secondmean{0.9753}\sd{0.0005} \\
B1 & ConvNeXt-Tiny & \secondmean{0.9736}\sd{0.0226} & \secondmean{0.9800}\sd{0.0164} & 0.9752\sd{0.0212} \\
B1 & ViT-B/16* & \bestmean{0.9837}\sd{0.0123} & \bestmean{0.9861}\sd{0.0087} & \bestmean{0.9842}\sd{0.0125} \\
\midrule
B2 & ResNet-50 & \bestmean{0.8778}\sd{0.0321} & \secondmean{0.8219}\sd{0.0227} & \secondmean{0.7878}\sd{0.0233} \\
B2 & ConvNeXt-Tiny* & \secondmean{0.8621}\sd{0.0317} & \bestmean{0.8421}\sd{0.0072} & \bestmean{0.8072}\sd{0.0324} \\
B2 & ViT-B/16 & 0.8231\sd{0.0279} & 0.8061\sd{0.0179} & 0.7816\sd{0.0272} \\
\bottomrule
\end{tabular}
\par\bigskip
\textbf{(b) Top-3 accuracy and probability quality}\par\smallskip
\begin{tabular}{@{}llccc@{}}
\toprule
Task & Backbone & Top-3 accuracy & ECE & Brier \\
\midrule
A & ResNet-50 & 0.9931\sd{0.0018} & 0.0332\sd{0.0031} & 0.0689\sd{0.0020} \\
A & ConvNeXt-Tiny & \bestmean{0.9989}\sd{0.0006} & \secondmean{0.0058}\sd{0.0028} & \bestmean{0.0137}\sd{0.0041} \\
A & ViT-B/16* & \secondmean{0.9984}\sd{0.0006} & \bestmean{0.0056}\sd{0.0014} & \secondmean{0.0165}\sd{0.0055} \\
\midrule
B1 & ResNet-50 & \secondmean{0.9984}\sd{0.0013} & 0.0198\sd{0.0027} & 0.0372\sd{0.0020} \\
B1 & ConvNeXt-Tiny & 0.9973\sd{0.0041} & \secondmean{0.0102}\sd{0.0078} & \secondmean{0.0309}\sd{0.0259} \\
B1 & ViT-B/16* & \bestmean{0.9988}\sd{0.0005} & \bestmean{0.0073}\sd{0.0018} & \bestmean{0.0209}\sd{0.0133} \\
\midrule
B2 & ResNet-50 & \bestmean{0.9559}\sd{0.0168} & \bestmean{0.0448}\sd{0.0163} & \secondmean{0.2581}\sd{0.0315} \\
B2 & ConvNeXt-Tiny* & \secondmean{0.9498}\sd{0.0169} & \secondmean{0.0831}\sd{0.0478} & \bestmean{0.2523}\sd{0.0312} \\
B2 & ViT-B/16 & 0.9456\sd{0.0026} & 0.1013\sd{0.0403} & 0.3050\sd{0.0419} \\
\bottomrule
\end{tabular}
\end{table}